\documentclass[preprint,12pt,times]{elsarticle}
\usepackage{amssymb}
\usepackage{amsmath}
\usepackage{graphicx,booktabs,multirow,makecell}
\usepackage{subfigure} 
\usepackage{algorithmic}
\usepackage{array}
\usepackage{algorithm}
\usepackage{utfsym}
\usepackage{xcolor}
\usepackage{multirow}
\journal{Nuclear Physics B}
\usepackage{hyperref}
\hypersetup{
	colorlinks=true,
	linkcolor=blue,  %
	citecolor=blue,  %
	urlcolor=blue,   
	linkbordercolor=0 0 0, 
}
\begin{document}
\begin{frontmatter}



\title{H-SGANet: Hybrid Sparse Graph Attention Network for Deformable Medical Image Registration}

\author[label1]{Yufeng Zhou} 
\ead{2250432015@email.szu.edu.cn}
\author[label1]{Wenming Cao\corref{cor1}} 
\ead{wmcao@szu.edu.cn}
\cortext[cor1]{Corresponding author}
\affiliation[label1]{organization={The State Key Laboratory of Radio Frequency Heterogeneous Integration},
            addressline={Shenzhen University},
            city={Shenzhen},
            postcode={518060},
            state={Guangdong Province},
            country={China}}

\begin{abstract}
The integration of Convolutional Neural Network (ConvNet) and Transformer has emerged as a strong candidate for image registration, leveraging the strengths of both models and a large parameter space. However, this hybrid model, treating brain MRI volumes as grid or sequence structures, faces challenges in accurately representing anatomical connectivity, diverse brain regions, and vital connections contributing to the brain's internal architecture. Concerns also arise regarding the computational expense and GPU memory usage associated with this model. To tackle these issues, a lightweight hybrid sparse graph attention network (H-SGANet) has been developed. This network incorporates a central mechanism, Sparse Graph Attention (SGA), based on a Vision Graph Neural Network (ViG) with predetermined anatomical connections. The SGA module expands the model's receptive field and seamlessly integrates into the network. To further amplify the advantages of the hybrid network, the Separable Self-Attention (SSA) is employed as an enhanced token mixer, integrated with depth-wise convolution to constitute SSAFormer. This strategic integration is designed to more effectively extract long-range dependencies. As a hybrid ConvNet-ViG-Transformer model, H-SGANet offers threefold benefits for volumetric medical image registration. It optimizes fixed and moving images concurrently through a hybrid feature fusion layer and an end-to-end learning framework. Compared to VoxelMorph, a model with a similar parameter count, H-SGANet demonstrates significant performance enhancements of 3.5\% and 1.5\% in Dice score on the OASIS dataset and LPBA40 dataset, respectively.
\end{abstract}



\begin{keyword}
Deformable image registration, deep learning, graph convolutional network, unsupervised learning, hybrid model.
\end{keyword}
\end{frontmatter}

\section{Introduction}
\label{sec:introduction}
Deformable medical image registration (DMIR) plays a crucial role in clinical applications such as diagnosis, surgery, therapy, and monitoring, holding significant theoretical and practical importance \cite{deng2023interpretable}. The goal of DMIR is to establish anatomical correspondence by predicting a deformation field that aligns one image (the moving image) with another (the fixed image). Conventional image registration methods, such as SyN \cite{avants2008symmetric} and NiftyReg \cite{modat2010fast}, maximize similarity and produce smooth deformation fields through iterative optimization of transformations. However, these methods are associated with considerable processing delays and high computational demands. With the rapid advancement of convolutional neural networks (ConvNets), learning-based deformable registration methods have gained prominence due to their marked superiority in registration accuracy and speed \cite{sotiras2013deformable,chen2023survey}. These deep learning-driven approaches achieve state-of-the-art accuracy and are capable of completing end-to-end registration within seconds. However, achieving accurate and efficient registration remains a formidable challenge, particularly with large deformations and volumetric data demands. This issue highlights the inherent trade-offs between different algorithmic focuses within network architectures; some may excel in accuracy while incurring higher model complexity. Considering the distribution of structures in brain space, such as the intricate arrangement of gyri and sulci, the spatial relationships between different brain regions, and the topological features that are critical for neurological function, it becomes evident that the brain's spatial architecture plays a pivotal role in anatomical correspondence \cite{li2023d}. To enhance efficiency while taking into account the inherent spatial architecture of the brain, there is a need for a hybrid model that leverages the anatomical connectivity of the brain.

Our prior research has exhibited initial findings indicating the promising performance of the integration between the hybrid model and V-Net \cite{milletari2016v} in the realm of image registration, which has demonstrated the effectiveness of this approach \cite{zheng2022multi}. The convergence and integration of various disciplines have led to the development of a multitude of DMIR schemes based on learning and other techniques, which are progressively emerging. However, despite the diversity of these approaches, they all rely upon the core principles of three key methodologies: ConvNets focus on local features within medical imagery, while Transformer-based methods emphasize the importance of multi-scale features. Additionally, Graph Convolutional Neural Networks (GCNs) offer a new perspective by highlighting spatial relationships within medical images. Although distinct in their implementation and focus, these three approaches collectively constitute the fundamental basis for contemporary DMIR solutions, each contributing a unique and valuable dimension to the understanding and manipulation of complex anatomical structures.

Due to the outstanding performance demonstrated by deep learning techniques, particularly the formidable capacity of ConvNets in feature extraction, a plethora of deep learning models have been leveraged for brain registration tasks, resulting in commendable outcomes. VoxelMorph is proposed in \cite{balakrishnan2019voxelmorph}, which features a mere 0.27M parameters and achieves superior performance through computational efficiency by replacing per-image optimization with global function optimization. However, ConvNet’s limited effective receptive fields due to locality nature hinder their performance in image registration, which often necessitates establishing long-range spatial correspondences between images. To address this, TransMorph \cite{chen2022transmorph}, a hybrid Transformer-ConvNet method, utilizes a U-Net architecture \cite{ronneberger2015u} with Swin Transformer \cite{liu2021swin} encoder and ConvNet decoder. In spite of its hybrid implementation, TransMorph introduces a considerable augmentation in the number of network parameters and computational complexity, thereby potentially undermining the strengths of the hybrid model. Particularly, the inclusion of multi-headed self-attention (MHA) in Transformers exacerbates these demands due to its ${O(k^2)}$ time complexity with respect to the number of tokens or patches $ k $. The combination of a large parameter count and intensive calculations poses challenges, thereby limiting the applicability of Transformer-based DMIR models.

This raises the question of how a hybrid model can fuse the strengths of various networks for DMIR. The vision graph neural network (ViG) \cite{han2022vision} is a very effective mechanism for capturing the intrinsic relationship of irregular objects because it foresees a graph structure that comes from the inherent connectivity and spatial layout of these objects. ViG divides the image into smaller, fixed-sized patches, facilitating the transformation of continuous image data into discrete, manageable units. By doing so, each patch contains information about a portion of the image, such as color, texture, and other pertinent features. Leveraging the $ K $-Nearest Neighbors (KNN) algorithm, ViG identifies and establishes connections between patches exhibiting similarity in features. These connections effectively capture the intricate relationships and interactions among distinct regions within the image. ViG is capable of capturing and processing global object interactions, thereby enabling a high level of performance in understanding the anatomical connectivity within the image. However, challenges persist in applying ViG to registration network. Computing KNN for each input image pair, given the unpredictable nature of nearest neighbors, leads to resource-intensive operations. The need to reshape the input image multiple times increases computational costs, particularly when deploying the model on a GPU. Furthermore, the KNN algorithm's generalization capabilities are limited, necessitating significant computational resources and complicating its use as a generalized module for DMIR.

In this study, we propose a hybrid sparse graph attention network (H-SGANet), a novel hybrid ConvNet-ViG-Transformer framework for volumetric medical image registration. Our framework features a Sparse Graph Attention (SGA) module designed to establish connections between nodes across diverse anatomical regions. By leveraging graph neural networks, the SGA module effectively formulates and aggregates structure-to-structure relationships, capturing the brain's inherent anatomical structure. Furthermore, the SGA module streamlines the KNN computation and obviates the need for input data reshaping, thereby enhancing computational efficiency without incurring additional computational overhead. Oriented on the Transformer, we have incorporated a Separable Self-Attention (SSA) mechanism as a token mixer, effectively replacing the traditional MHA component within the Transformer model. This innovative adaptation replaces batch-wise matrix multiplication with element-wise operations, significantly reducing the computational burden. Specifically, it transforms the time complexity from the quadratic ${O(k^2)}$ associated with MHA to a linear complexity, without compromising the hybrid architecture's ability to incorporate high-quality long-range dependencies. This modification not only maintains the efficiency of the model but also prevents a concomitant increase in computational complexity.

In summary, we present the following key contributions:
\begin{itemize}
	\item To address the stereotypical treatment of images, we introduce Sparse Graph Attention (SGA). By combining the advantage of ViG, SGA enlarges receptive fields, facilitating flexible connectivity between specific internal brain regions and aiding in the identification of anatomical connectivity. Additionally, SGA does not require reshaping and imposes minimal computational overhead, distinguishing it from typical ViG-based methods.
	\item To effectively capture long-range dependencies and address the efficiency bottleneck in MHA, we introduce the SSAFormer module, combining depthwise convolutional scaling (DCS) and Separable Self-Attention (SSA). Leveraging the scalability and robustness to corruption inherent in Transformers, the SSAFormer module enhances the feature extraction capability of the hybrid model while concurrently minimizing computational costs.
	\item We present H-SGANet, a lightweight hybrid architecture that combines ConvNet-ViG-Transformer by integrating SGA and SSAFormer within a U-shaped framework. With a parameter count of 0.382M, H-SGANet surpasses existing state-of-the-art baselines in terms of accuracy, thereby showcasing its potential in the domain of DMIR.
\end{itemize}

\section{RELATED WORK}
\subsection{ViT-based deformable registration methods}
Vision Transformer (ViT) \cite{dosovitskiy2020image} models have gained prominence in 3D medical image registration. Well-known earliest hybrid model ViT-V-Net \cite{chen2021vit}, which merges the strengths of ViT with the established V-Net architecture, significantly enhances the capabilities of established models like VoxelMorph. A dual transformer network (DTN) is proposed in \cite{zhang2021learning}, which uses two identical encoder branches to model the cross-volume dependencies. These approaches employ conventional self-attention to forge relationships between concatenated tokens derived from moving and fixed images. As mentioned earlier, TransMorph \cite{chen2022transmorph} substitutes the encoder of ViT-V-Net with powerful Swin blocks \cite{liu2021swin}, which utilize a window-based self-attention method operating within local windows to compute attention for concatenated tokens. These methods enhance the understanding of spatial correspondence within and between images. However, they fall short in capturing correlations across moving and fixed images, limiting the ability of hybrid model for fine registration. Recent studies have sought to augment the self-attention mechanism with cross-attention to better comprehend spatial correspondences. For instance, Cross-modal attention \cite{SONG2022102612} employs conventional cross-attention to establish relationships between fixed and moving image tokens. In contrast, Xmorpher \cite{shi2022xmorpher} operates within two distinct local windows of varying sizes to compute attention between corresponding tokens, enhancing the registration model's ability to exchange information through a cross-attention mechanism. The global computation of cross-attention restricts hierarchical feature extraction and is primarily suited for lower-scale features. To enhance feature extraction and matching, TransMatch \cite{10158729} integrates local window self-attention and cross-attention within a dual-stream framework. However, memory and processing speed become increasingly difficult as computational complexity increases. In other words, despite the great potential shown by the hybrid mentioned above ViT-based registration methods, the utilization of spatial correspondences for DMIR is still in its infancy and many challenges remain.

\subsection{Attention mechanism and Vision GNN (ViG)}
Aiming to enhance model performance with minimal computational overhead, a variety of attention mechanisms have been integrated into network models. These include Coordinate Attention (CA) \cite{hou2021coordinate}, Efficient Channel Attention (ECA) \cite{zhang2021efficient}, and Separable Self-attention (SSA) \cite{mehta2022separable}. Prior to the advent of ViT-V-Net \cite{chen2021vit}, self-attention mechanisms were introduced for sequence-to-sequence prediction in medical image segmentation by \cite{chen2021transunet}. The VAN network \cite{zu2021van}, which builds upon the Convolutional Block Attention Module (CBAM) \cite{woo2018cbam}, has improved registration performance through a voting process that refines the final deformation field. Furthermore, \cite{chen2023dusfe} introduced the Dual-Channel Squeeze-Fusion-Excitation (DuSFE) co-attention method, which enhances the performance of SE blocks \cite{hu2018squeeze} by employing a co-attention mechanism for gradual feature fusion in cross-modality image registration. Despite these methods' informative feature extraction capabilities, they may fall short in capturing the intricate spatial relationships within graph-structured medical data, which are crucial for accurate medical image registration.

Since their extensive study beginning with \cite{gori2005new,scarselli2008graph}, Graph Neural Networks (GNNs) have evolved with various Graph Convolutional Network (GCN) variants  \cite{niepert2016learning,kipf2016semi,defferrard2016convolutional} and found applications in diverse fields  \cite{hamilton2017inductive,ying2018graph,kojima2020kgcn}. Recently, the Vision Graph Neural Network (ViG) \cite{han2022vision} has demonstrated comparable performance to ViT models like DeiT \cite{touvron2021training} and Swin Transformer \cite{liu2021swin} in the domain of computer vision tasks. ViG represents images as graph structures, connecting image patches as nodes based on their nearest neighbors, facilitating global object interaction processing. Methods such as GraformerDIR \cite{yang2022graformerdir} and DieGraph \cite{wang2023diegraph} have enhanced DMIR by integrating ViG, leveraging graph convolution to capture long-range dependencies and expanding the receptive field.

However, ViG's reliance on the $K$-nearest neighbors (KNN) algorithm for establishing patch connections, while effective, is computationally demanding, especially with a large number of features. Additionally, the process of reshaping the input from an $n$-dimensional tensor to ($n-1$)-dimensional for graph convolution and then restoring it to its original state is also time-consuming. To address these challenges, we introduce a sparse graph attention mechanism module designed to integrate seamlessly into various models. This module enhances the understanding of spatial architecture by focusing on the most salient connections within the brain's network, thereby effectively capturing the fixed connectivity relationships inherent in medical images and improving the precision of DMIR.
\begin{figure}[!ht]
	\centering \includegraphics[width=\linewidth, keepaspectratio]{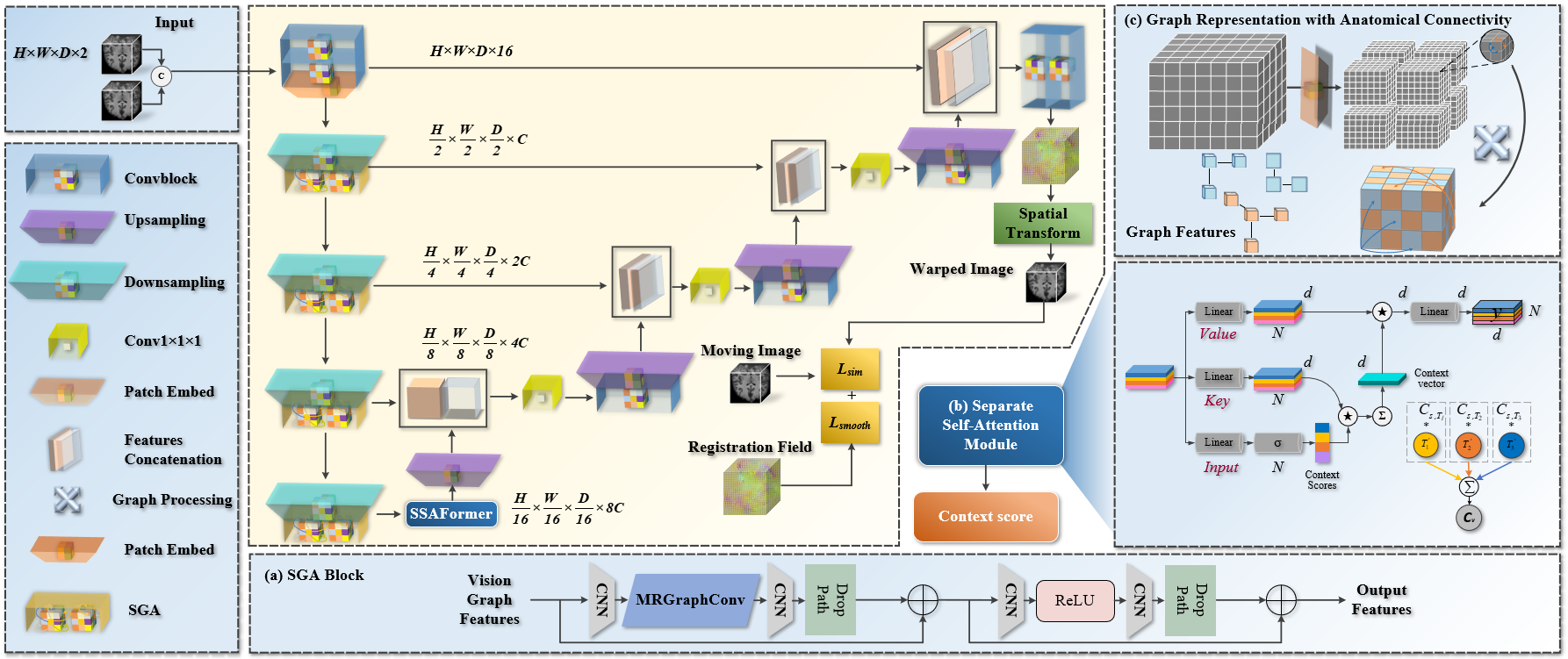}
	\caption{Overall block diagram of H-SGANet. It primarily features two core modules: (a) The SGA block leverages the graph representation depicted in block (c) to learn both global and graph-level representations. (b) The SSAFormer, contrasting with MHA's attention matrix, generates context scores and vectors using efficient element-wise operations, optimized for hybrid models processing volumetric data.}\label{fig1}
\end{figure}

\section{METHODS}
\subsection{Overall of network}
Our proposed H-SGANet is built on a U-Net architecture and incorporates two additional components: the Sparse Graph Attention (SGA) module and the Separable Self-Attention Former (SSAFormer). As shown in Fig. \ref{fig1}, H-SGANet takes a pair of images as input. Initially, features are extracted from the moving and fixed volumes using an encoder with a graph structure. The encoder includes an SGA module (c.f., Section 3.2) and a downsampling module. The graph structure splits the extracted feature maps into multiple patches and constructs the graph through a sparse rolling strategy (c.f., Algorithm 1). This approach enhances spatial correspondences using anatomical connectivity. As the encoding stages advance, the feature maps from earlier SGA blocks progressively represent the image as a graph (as shown in Fig. \ref{fig1}(c)). This representation captures global relationships and interactions among different regions, offering a more 'graph-level' understanding of the volumetric data. Subsequently, the features organized into graph structures are processed by the SSAFormer (c.f., Section 3.3) for extracting long-range dependencies (as shown in Fig. \ref{fig1}(b)). Positioned at the bottom of the network, the SSAFormer is responsible for capturing long-range dependencies, thereby further enhancing the performance of the network. Skip connections fuse features from different resolution levels, improving the accuracy and consistency of the deformation field. After four upsampling stages in the decoder, the network aligns the features of the two input images to create an alignment field. A spatial transformation function is then applied to align image $M$ with image $F$. The registration process is refined using a similarity loss function that compares the aligned images, supplemented by a smoothness loss term to regularize the registration field.

The purpose of unsupervised non-rigid image registration is to estimate a dense deformation field $\Tilde{\phi}:\mathbb{R}^{n}\rightarrow \mathbb{R}^{n}$, where $n$ represents the size of a 3D image, calculated as $n=H\times W\times D$, with $H$, $W$, and $D$ denoting the image height, width, and depth dimensions, respectively. Given a moving image $M$ and a fixed image $F$, image registration is framed as an optimization problem where the goal is to minimize the dissimilarity metric $\mathcal{D}$ (which maintains intensity constancy) and a regularization penalty $\mathcal{R}$ (which enforces smoothness):
\begin{equation}
\label{eq5}
\Tilde{\phi}=\mathop{\arg\min}\limits_{\phi}\left(\mathcal{D}\left(\phi\left(M\right),F\right)+\lambda\mathcal{R}\left(\phi\right)\right)
\end{equation}
where $\Tilde{\phi}$ denotes the optimal estimated deformation field, and $\lambda$ a hyperparameter that controls the trade-off between the two optimization terms. Recent studies typically assume that affine preprocessing has been applied to correct global misalignment, focusing primarily on non-linear local deformations. 

\subsection{Sparse Graph Attention (SGA)}
\begin{figure}[htpb] 
	\centering \includegraphics[width=0.8\linewidth]{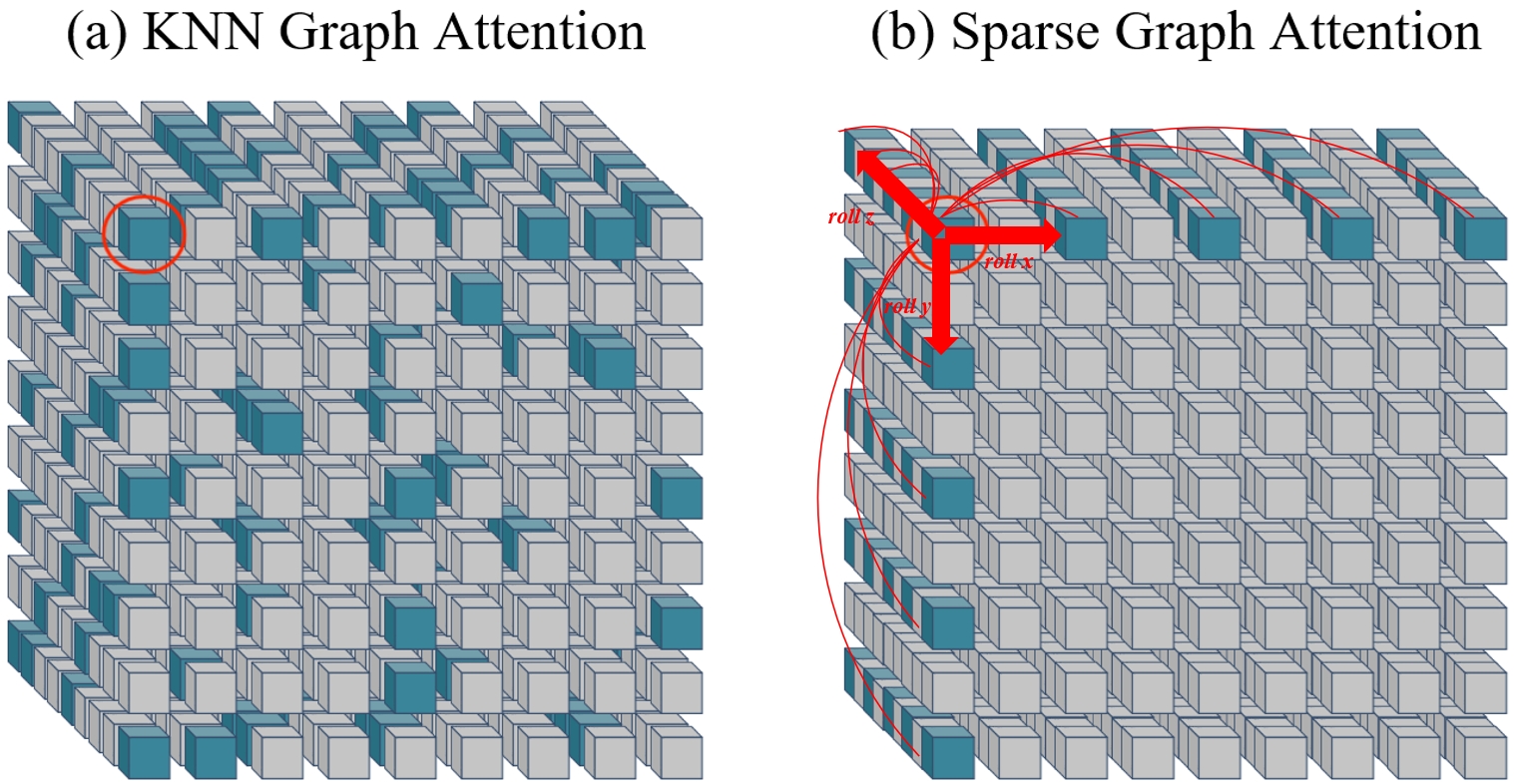}
	\caption{(a) In ViG, KNN graph attention allows the corner voxel of a 9×9×9 patch to form relationships with other voxels via random sampling. (b) SGA at the corresponding voxel position. In contrast to (a), SGA utilizes a structured graph, thereby eliminating the computationally intensive requirements of KNN and the need for data reshaping.}
	\label{fig2}	
\end{figure}

It is well-known that ViG incurs a significant computational load, primarily due to the KNN algorithm's need to process each input image patch. Consequently, this creates a structural pattern characterized by seemingly random connections, as shown in Fig. \ref{fig2}(a). Additionally, the unstructured nature of KNN poses another challenge: the reshaping operations are computationally expensive. The input image must be transformed from a 4D to a 3D tensor to align connected pixel features for graph convolution. After the graph convolution, the 3D tensor is converted back to a 4D format for the following convolutional layers. Thus, KNN-based attention involves computationally intensive KNN operations and two expensive reshaping steps, which present significant challenges in applying to medical images. This allows for the development of a simple yet effective graph attention module that aligns with the brain's inherent anatomical structure.

To reduce the computational overhead of KNN calculations and reshaping operations, a fixed anatomical connectivity graph structure is used, where each voxel is connected to every $K$th voxel in its row, column, and depth directions. For example, in a 9×9×9 patch with $K = 2$, a voxel at the top corner connects to every second voxel along its row, column, and depth directions, as illustrated in Fig. \ref{fig2}(b). This fixed connectivity pattern is uniform across all voxels in the input volume. As the graph structure is consistent (i.e., each voxel maintains the same connections across all patches), reshaping the input volume for graph convolution is unnecessary.

Specifically, the scrolling operations \textit{roll}$_{\text{x}}$, \textit{roll}$_{\text{y}}$, and \textit{roll}$_{\text{z}}$ are applied across the three voxel dimensions, as outlined in Algorithm \ref{alg:SGA}. These scrolling operations are controlled by two key parameters: the first specifies the input voxel for the rolling process, and the second defines the magnitude and direction of the scroll in the width, height, and depth dimensions. These settings are crucial for modifying input data and acquiring graph-level information from neighboring locations. For example, in Fig. \ref{fig2}(b), the voxel at the top-right corner of the patch can be aligned with every second voxel along the $X$ dimension by rolling two units to the right. A similar process can be applied to the $y$ and $z$ dimensions by rolling in different directions. SGA adopts max-relative graph convolution \cite{li2019deepgcns} to aggregate neighborhood relative features. Consequently, after each \textit{roll}$_{\text{x }}$, \textit{roll}$_{\text{y }}$, and \textit{roll}$_{\text{z }}$ operation, it is necessary to compute the difference between the original input voxel and its rolled version, represented as \textit{V}$_{\text{x}}$, \textit{V}$_{\text{y}}$, and \textit{V}$_{\text{z}}$, respectively, to capture the relative features along each dimension. The final $\text{Conv3d}$ operation is performed after completing the rolling and max-relative operations (MRConv). This method allows SGA to use rolling operations to construct a simpler and more cost-effective graph. 
\begin{algorithm}[!ht]
	\caption{3D-SGA with MRConv}
	\label{alg:SGA}
	\begin{algorithmic}[1]
		\REQUIRE $K$ represents the distance between connections, while $H$, $W$, and $D$ denote the image resolution in three dimensions. Here, $X$ refers to the 3D input image, and $m$ governs the distance of each roll
		\ENSURE Conv3d(Concat($X$, $X_j$))
		\STATE $m \gets 0$
		\WHILE{$mK < H$}
		\STATE $X_c \gets X - \text{roll}_{x}(X, mK)$ \COMMENT{get relative features along the depth dimension}
		\STATE $X_j \gets \max(X_c, X_j)$ \COMMENT{keep max relative features along the depth dimension}
		\STATE $m \gets m + 1$
		\ENDWHILE
		
		\STATE $m \gets 0$
		\WHILE{$mK < W$}
		\STATE $X_r \gets X - \text{roll}_{y}(X, mK)$ \COMMENT{get relative features along the width dimension}
		\STATE $X_j \gets \max(X_r, X_j)$ \COMMENT{keep max relative features along the width dimension}
		\STATE $m \gets m + 1$
		\ENDWHILE
		\STATE $m \gets 0$
		\WHILE{$mK < D$}
		\STATE $X_d \gets X - \text{roll}_{z}(X, mK)$ \COMMENT{get relative features along the depth dimension}
		\STATE $X_j \gets \max(X_d, X_j)$ \COMMENT{keep max relative features along the depth dimension}
		\STATE $m \gets m + 1$
		\ENDWHILE
		
		\RETURN Conv3d(Concat($X$, $X_j$))
	\end{algorithmic}
\end{algorithm}

In deep GCNs, when nodes become difficult to distinguish due to over-smoothing, the rapid information transfer between neighboring nodes degrades performance in computer vision tasks. In contrast, diverse node features enhance node distinction. To enhance node feature diversity, we introduce additional feature transformations and nonlinear activation functions based on SGA with MRConv. To project node features into the same domain and increase feature variety, we apply a linear layer before and after the optimized Max-Relative GCN, replacing the Grapher block proposed in Vision GNN. To prevent layer collapse, a nonlinear activation function is used after the optimized Max-Relative GCN. For the input feature $X \in \mathbb{R}^{N \times D}$, the revised Grapher is formulated as:
\begin{equation}
Y=\sigma\left(M R \operatorname{Conv}\left(X W_{\text {in }}\right)\right) W_{\text {out }}+X,
\label{eq1}
\end{equation}
where $Y \in \mathbb{R}^{N \times D}$, $W_{in}$ and $W_{out}$ compose the weights for fully connected layers, $\sigma$ is a GeLU activation.  
A two-layer MLP, referred to as a feed-forward network (FFN), is employed for graph processing to ensure effective feature translation and address potential issues related to over-smoothing:
\begin{equation}
Z=\sigma\left(X W_1\right) W_2+Y,
\label{eq2}
\end{equation}
where $Z \in \mathbb{R}^{N \times D}$, the weights $W_{1}$ and $W_{2}$ pertain to the fully connected layer, and $\sigma$ is a GeLU activation. Batch normalization, which is omitted in Eq.\ref{eq1} and \ref{eq2} for concision, is applied after every fully-connected layer or graph convolution layer in the Grapher and FFN modules.

\subsection{Separable Self-attention in metaformer (SSAFormer)}
\begin{figure}[!ht] 
	\centering  
	\vspace{-0.25cm} 
	\includegraphics[width=0.87\linewidth]{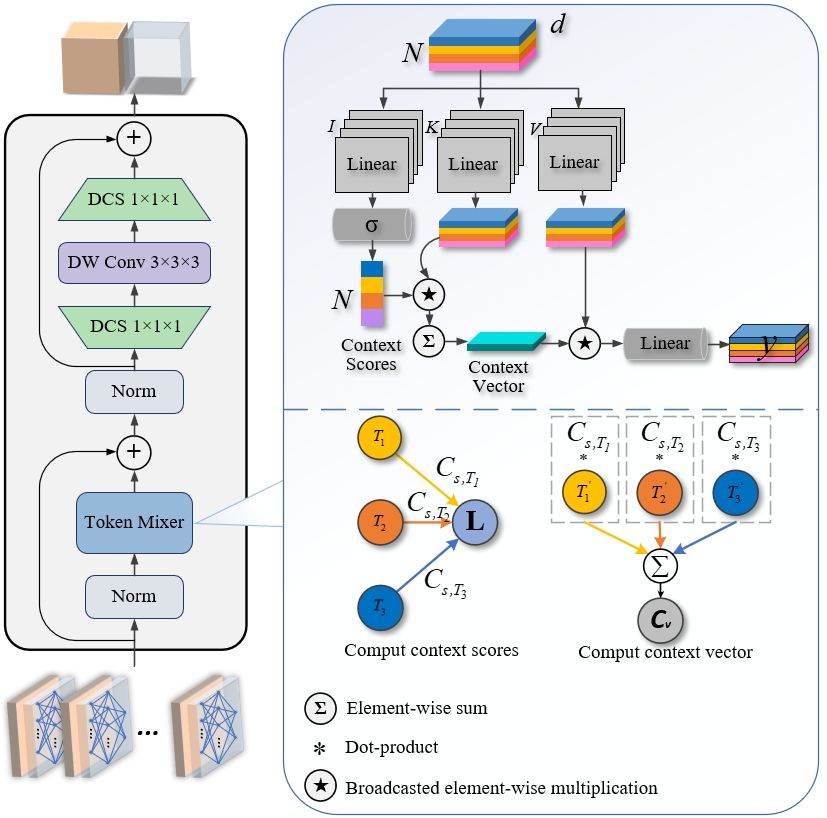}
	\caption{Overall block diagram of SSAFormer. Depthwise convolution and pointwise scaling are used as channel MLP to capture texture information. Token Mixer calculates context scores for each token with respect to a latent token $L$, leading to the generation of a context vector $c_v$ that captures global context with linear complexity ${O(k)}$} 
	\label{fig3}
\end{figure}

Recent years have witnessed great progress in Vision Transformer, which has been recognized and has long been investigated. MetaFormer \cite{yu2022metaformer} is a generic architecture abstracted from Transformer, which does not designate particular token mixer types. Specifically, it comprises two essential residual components: a token mixer, promoting information exchange among various tokens, and MLP blocks, tasked with feature extraction within the channel dimension. Notably, PoolFormer \cite{yu2022metaformer} surpasses ResNet \cite{he2016deep} using only pooling layers instead of self-attention, underscoring the paramount importance of the general Transformer structure. For efficient alignment of 3D medical images, based on MetaFormer structure is proposed in this paper, as shown in Fig. \ref{fig3}.

The Separable Self-attention mechanism is employed as the token mixer, calculating context scores concerning a latent token $L$ and using these scores to re-weight the input tokens, thereby generating a context vector that captures global information as depicted in Fig. \ref{fig3}. After the preceding process, the feature maps are obtained and then are linearly projected to generate the input tokens $H_i$ for the SSAFormer model. Each token is mapped through weight matrices corresponding to three branches: input ($\mathcal{I}$), key ($\mathcal{K}$), and value ($\mathcal{V}$). Within the input branch ($\mathcal{I}$), each d-dimensional token in $H_i$ is transformed into a scalar using a linear layer with weights $\mathbf{W}_{\mathbf{I}} \in \mathbb{R}^d$, serving as the latent node $L$:
\begin{equation}
L_i=W_I\cdot H_i,
\label{eq3}
\end{equation}
where $W_{I}$ is the weight matrix for the input branch ($\mathcal{I}$). The distance between token $L$ and $H_i$, referred to as context scores $\mathbf{c}_{\mathbf{s}} \in \mathbb{R}^k$, is computed by employing an inner-product operation:
\begin{equation}
\mathrm{Score}(H_i,L)=\frac{H_i^TL_i}{\sqrt{d}},
\label{eqadd3}
\end{equation}
where ${d}$ is the dimension of the embedding vector. These scores are subsequently utilized for computing the context vector $c_v$. In particular, the input sequence $H_i$ undergoes linear projection into a d-dimensional space through the key branch $K$, employing weights $\mathbf{W}_{\mathbf{K}} \in \mathbb{R}^{d \times d}$. The context vector $\mathbf{c}_{\mathbf{v}} \in \mathbb{R}^d$ is obtained by weighted summation calculation:
\begin{equation}
c_v=\sum_{i=1}^kc_{s_i}\cdot(W_K\cdot H_i),
\label{eq4}
\end{equation}
where $W_K$ is the weight matrix for the key branch ($\mathcal{K}$).
Compared to traditional attention mechanisms (${\mathcal{O}}(k^2)$) complexity, SSA effectively lowers the computational expense of calculating attention scores to ${\mathcal{O}}(k^{})$, making it computationally more efficient.

Then, the contextual information is efficiently propagated to each token in the input sequence ($H_i$) through broadcasted element-wise multiplication, as expressed by:
\begin{equation}
{\mathrm{propagated\_context}}=c_v\odot{\mathrm{ReLU}}(W_V\cdot H_i),
\label{eq5}
\end{equation}
where $W_V$ is the weight matrix for the value branch ($\mathcal{V}$), $\odot$ represents the element-wise multiplication, and $propagated_context$ captures the shared context information efficiently distributed to all tokens in $H_i$. This process ensures that every token in the sequence benefits from the same shared context vector ($\mathbf{c}_{\mathbf{v}}$), enhancing the model's ability to capture and utilize relevant information. The output of SSAFormer is as follow:
\begin{equation}
y=W_O\cdot\text{propagated}\_\text{context},
\label{eq6}
\end{equation}
where $\mathrm{W}_{\mathrm{O}}\in\mathbb{R}^{d\times\bar{d}}$ is the weight matrix for the linear transformation of the final output.

\subsection{Loss Function}
The network training uses an unsupervised learning approach with a loss function containing two parts: similarity loss $L_{sim}$ measuring the similarity between fixed and warped images, and a regularization term smoothing the predicted deformation field. 
\begin{equation}
L(F,M,\phi)=L_{sim}(F,M\circ\phi)+\lambda L_{reg}(\phi),
\label{eq7}
\end{equation}
where $\lambda $ is a regularization trade-off parameter, generally set to 1.

\subsubsection{Similarity Loss}
In DMIR, similarity metrics assess the closeness of medical images. Maximizing these metrics optimizes image alignment. In our proposed network, we adopt two widely used similarity measures in the loss function. One is the Mean Squared Error(MSE), which calculates the squared intensity differences between corresponding voxels of the warped moving image $M\circ\phi $ and the fixed image $F$:
\begin{equation} \label{e8}
\begin{aligned}
MSE(F, M \circ \phi) = \frac{1}{| \Omega |} \sum_{p \in \Omega} |F(p) - M \circ \phi(p)|^2,
\end{aligned}
\end{equation}
where $\Omega$ denotes the image domain, and $p$ refers to the voxel position in the image domain.

A second one is the local normalized cross-correlation (LCNN), defined as:
\begin{equation} \label{e11}
	{\small
		LNCC(F, M \circ \phi)=
		\sum_{\substack{\mathbf{p}\in\Omega  }}\frac{\left(\sum_{\mathbf{p}_i} (F(\mathbf{p}_i) - \bar{f}(\mathbf{p}))(M \circ \phi(\mathbf{p}_i) - \bar{M}{ \circ \phi}(\mathbf{p}))\right)}{\sqrt{\left(\sum_{\mathbf{p}_i} (F(\mathbf{p}_i) - \bar{f}(\mathbf{p}))^2\right)}\sqrt{\left(\sum_{\mathbf{p}_i} (M \circ \phi(\mathbf{p}_i) - \bar{M}{ \circ \phi}(\mathbf{p}))^2\right)}}}.
\end{equation}
where $F(\mathbf{p}_i)$ denotes the denote the intensity of the fixed image at voxel location $\mathbf{p}_i$, and $\bar{f}(\mathbf{p})$ represent the mean intensity within a local $n^3$-sized window centered at voxel $\mathbf{p}$. Similarly, $M \circ \phi(\mathbf{p}_i)$ represents the intensity of the warped moving image after registration, with $\bar{M} \circ \phi(\mathbf{p})$ denoting the mean intensity of the warped moving image within the local window.
Finally, the similarity loss function is defined as:
\begin{equation} \label{e12}
\begin{aligned}
L_{sim} = 1 - LNCC(F, M \circ \phi).
\end{aligned}
\end{equation}

\subsubsection{Regularization Loss of Deformation Field}
We leverage deformation field gradients to penalize abrupt changes and foster a smoother field. This regularization significantly mitigates unrealistic deformations in medical image registration, enhancing the algorithm's robustness and reliability. Specifically, we employ the L2 norm on spatial gradients:
\begin{equation} \label{e13}
\begin{aligned}
L_{reg}(\phi) = \sum_{p \in \Omega} || \nabla u(p) ||^2,
\end{aligned}
\end{equation}
where $\nabla u$ denotes the spatial gradient of the velocity field $u$.

\section{Experiments}
In this section, we validate the performance of the proposed H-SGANet on OASIS and LPBA40 datasets for DMIR experiments and compare the results with recent state-of-the-art DMIR methods. For neural network training, all models are trained from scratch, without pretraining. To verify the effectiveness of major components, we also perform a series of ablation studies and comparison experiments with other attention modules.

\subsection{Data and Pre-processing}
\begin{table}[htpb]
	\centering
	\caption{The 28 anatomical structures were used to test in learn2reg OASIS brain MRI registration}
	\setlength{\tabcolsep}{1.5pt}
	\small
	\begin{tabular}{@{}cccc@{}}
		\toprule
		Label & Structure Name & Label & Structure Name  \\
		\midrule
		2 & Left-cerebral-white-matter & 46 & Right-cerebral-white-matter \\
		3 & Left-cerebral-cortex & 42 & Right-cerebral-cortex \\
		4 & Left-lateral-ventricle & 43 & Right-Lateral-Ventricle \\
		7 & Left-cerebellum-white-matter & 48 & Right-cerebellum-white-matter \\
		8 & Left-cerebellum-cortex & 47 & Right-cerebellum-cortex\\
		10 & Left-thalamus & 49 & Right-thalamus\\
		11 & Left-caudate & 50 & Right-caudate \\
		12 & Left-putamen & 51 & Right-putamen\\
		13 & Left-pallidum & 52 & Right-pallidum\\
		14 & 3rd-ventricle & 15 & 4th-ventricle\\
		17 & Left-hippocampus & 53 & Right-hippocampus \\
		18 & Left-amygdala & 54 & Right-amygdala\\
		16 & Brain-stem & 24 & CSF\\
		28 & Left-ventral-DC & 60 & Right-ventral-DC\\
		\bottomrule
	\end{tabular}
	\label{tab:1}%
\end{table}%
\begin{table}[!ht]
	\centering
	\caption{The 56 anatomical structures in LPBA40 were used for cross-dataset validation}
	\setlength{\tabcolsep}{1.5pt}
	\small
	\begin{tabular}{@{}cccc@{}}
		\toprule
		Label & Structure Name & Label & Structure Name  \\
		\midrule
		21 & L superior frontal gyrus & 65 & L inferior occipital gyrus \\
		22 & R superior frontal gyrus & 66 & R inferior occipital gyrus \\
		23 & L middle frontal gyrus & 67 & L cuneus \\
		24 & R middle frontal gyrus & 68 & R cuneus \\
		25 & L inferior frontal gyrus & 81 & L superior temporal gyrus\\
		26 & R inferior frontal gyrus & 82 & R superior temporal gyrus\\
		27 & L precentral gyrus & 83 & L middle temporal gyrus \\
		28 & R precentral gyrus & 84 & R middle temporal gyrus\\
		29 & L middle orbitofrontal gyrus & 85 & L inferior temporal gyrus\\
		30 & R middle orbitofrontal gyrus & 86 & R inferior temporal gyrus\\
		31 & L lateral orbitofrontal gyrus & 87 & L parahippocampal gyrus \\
		32 & R lateral orbitofrontal gyrus & 88 & R parahippocampal gyrus\\
		33 & L gyrus rectus & 89 & L lingual gyrus\\
		34 & R gyrus rectus & 90 & R lingual gyrus\\
		41 & L postcentral gyrus & 91 & L fusiform gyrus\\
		42 & L R postcentral gyrus & 92 & R fusiform gyrus \\
		43 & L superior parietal gyrus & 101 & L insular cortex\\
		44 & R superior parietal gyrus & 102 & R insular cortex\\
		45 & L supramarginal gyrus & 121 & L cingulate gyrus\\
		46 & R supramarginal gyrus & 122 & R cingulate gyrus \\
		47 & L angular gyrus & 161 & L caudate\\
		48 & R angular gyrus & 162 & R caudate\\
		49 & L precuneus & 163 & L putamen\\
		50 & R precuneus & 164 & R putamen \\
		61 & L superior occipital gyrus & 165 & L hippocampus\\
		62 & R superior occipital gyrus & 166 & R hippocampus\\
		63 & L middle occipital gyrus & 181 & Cerebellum\\
		64 & R middle occipital gyrus & 182 & Brainstem\\
		\bottomrule
	\end{tabular}
	\label{tab:2}%
\end{table}

The OASIS and LPBA40 datasets, which have more brain MR scans with segmentation of subcortical structures, are used in our experiments, and the LPBA40 is selected for cross-dateset validation. In OASIS dataset, there are 425 T1-weighted brain MR scans from 416 subjects aged 18 to 96, including 100 clinically diagnosed with mild to moderate Alzheimer's disease. To fully utilize the anatomical connectivity, a series of standard preprocessing steps, such as skull stripping, spatial normalization, and subcutaneous structure segmentation using FreeSurfer, are applied. Our primary emphasis lies in subcortical segmentation, involving 28 anatomical structures as the basis for evaluation as seen in Table \ref{tab:1}. Subsequently, for efficiency, all scans are uniformly cropped to 160×192×224 dimensions. The dataset is partitioned into 255, 20, and 150 volumes for training, validation, and testing. For network training, we use a subject-to-subject registration approach, which yields 64770 training pairings in total. During testing, five volumes are randomly selected as fixed atlases and the remaining volumes are registered as moving images to ensure robust evaluation so that there are 725 pairs.

The LPBA40 dataset comprises 40 brain MR scans, with evaluations performed using expert-drawn segmentations of 56 subcortical structures (as shown in Table \ref{tab:2}). MR scans from the LPBA40 dataset undergo the same preprocessing steps as those applied to the OASIS dataset. Notably, all volumes in this dataset are reserved exclusively for cross-dataset validation, which has 175 test pairs, not for training.

\subsection{Evaluating Indicator}
The performance of our proposed H-SGANet and other baseline methods was evaluated based on anatomical overlaps and deformation smoothness. Anatomical overlap was quantified using the Dice score (DSC), a widely accepted metric for assessing the alignment of anatomical structures in the images. Our datasets provide precise segmentation annotations, which are used solely for calculating DSC values between the fixed and deformed moving masks and do not contribute to the training of the model. Deformation smoothness was assessed using the non-normal Jacobian determinant (NJD). The Dice score measures volume overlap in anatomical and organ segmentations, providing a robust indicator of registration performance. For a given pair of images $F$ and $M$, along with their registration fields $\phi$, the Dice score is defined as followed:
\begin{equation}
\operatorname{Dice}(F, M, \phi)=\frac{1}{K} \sum_{k=1}^K \frac{2 \cdot\left|s_f^k \cap s_m^k\right|}{\left|s_f^k \cup s_m^k\right|}, 
\label{eq14}
\end{equation}
where $s_f^k$ and $s_m^k$ represent the $k$-th segmentation of K anatomical structures in images $F$ and $M$, respectively ($k \in[1, K]$). The Dice score measures the similarity of image pair, which falls with in the range of $[0,1]$.

The percentage of non-normal Jacobian determinant($\left|J_\phi\right| \leq 0$) on the deformation fields is employed as an evaluation metric, which is computed based on the coordinates of position p and serves as an indicator of the deformation field's impact on voxel changes. The Jacobian determinant is as follows:
\begin{equation}
\left|J_\phi\right| \leq 0=\sum_{p \in \Omega}\left(J_\phi(p) \leq 0\right),
\label{eq15}
\end{equation}
where 
\begin{equation}
J_\phi(p)=\begin{aligned}
\frac{\partial \phi_x(p)}{\partial x} &~~ \frac{\partial \phi_y(p)}{\partial x} ~~ \frac{\partial \phi_z(p)}{\partial x} \\
\frac{\partial \phi_x(p)}{\partial y} &~~ \frac{\partial \phi_y(p)}{\partial y} ~~ \frac{\partial \phi_z(p)}{\partial y} \\
\frac{\partial \phi_x(p)}{\partial z} &~~ \frac{\partial \phi_y(p)}{\partial z} ~~ \frac{\partial \phi_z(p)}{\partial z},
\label{eq16}
\end{aligned}
\end{equation}
$\left(\phi_x(p), \phi_y(p), \phi_z(p)\right)$ is the position p coordinates. A Jacobian determinant of 1 indicates no change in the voxel, values between 0 and 1 denote shrinkage, and values greater than 1 denote expansion. A determinant less than or equal to 0 implies folding, which is impractical and should be avoided. By comparing the percentage of voxels with non-normal Jacobian determinants, we can effectively evaluate the authenticity and smoothness of the registration field. This comprehensive approach ensures a more nuanced understanding of both anatomical overlaps and the quality of deformation fields in the evaluation process. 

Furthermore, we also consider multiple performance metrics to comprehensively assess the efficiency and resource utilization of our proposed method. First, the average running time is measured, providing insights into the speed of registering each pair of scans on the OASIS and LPBA40 datasets. Second, GPU memory usage is evaluated, indicating the memory footprint of our algorithm during the registration process. Third, we explore the model complexity through the assessment of parameters amount, shedding light on the computational demands of our approach. Lastly, the computational efficiency is quantified in terms of Giga Multiply-Accumulate Operations (GMACs), capturing the number of multiply-accumulate operations per second. These four metrics collectively contribute to a thorough understanding of our method's performance across different dimensions.

\subsection{Baseline Methods}	
To validate the superiority of H-SGANet, comparative experiments are conducted with various prominent works. This analysis encompasses two traditional registration algorithms, namely SyN \cite{avants2008symmetric} and Elastic \cite{bajcsy1989multiresolution}. Additionally, two fully convolutional networks, VoxelMorph \cite{balakrishnan2019voxelmorph} and LapIRN \cite{mok2020large}, are incorporated. For the Transformer-based methods, Vit-V-Net \cite{chen2021vit} and TransMorph \cite{chen2022transmorph} are added for comparison. For the four aforementioned deep learning-based methods, publicly accessible implementations are utilized, and strict adherence to the hyperparameter configurations outlined in their respective original publications is ensured during the training phase.

\subsection{Implementation Details}
Our proposed DMIR framework H-SGANet is implemented by PyTorch 1.10.0, and trained for 508 epochs on an NVIDIA RTX 3090 GPU. The network parameters are optimized using the Adam optimizer with a learning rate $1 \times 10^{-4}$ and a batch size of 1. The models from all epochs are then evaluated on the validation set, and the one with the best performance is chosen as the final prediction model.

\section{Result}
\begin{figure}[!ht]
	\centering
	\includegraphics[width=1\textwidth]{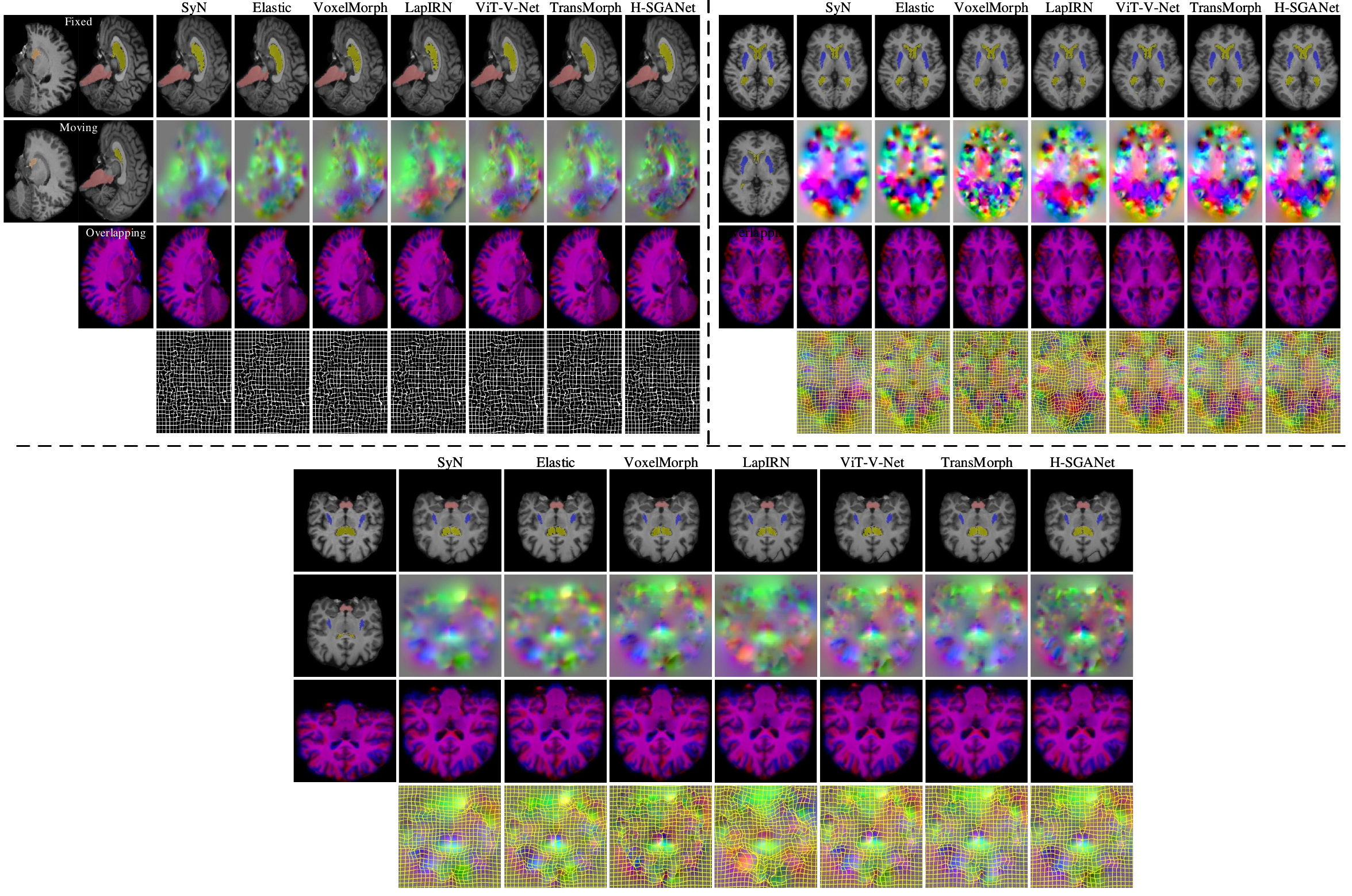}
	\caption{Qualitative comparison of various registration methods on the OASIS dataset. Lateral ventricles, putamen, and brain stem are color-coded yellow, blue, and red, respectively. Top-left, top-right and bottom panels display results on sagittal, coronal, and axial slices. In each panel, the first column exhibits the fixed image, moving image, and their overlap (a larger purple area indicates more overlap). The second and fourth rows represent the deformation field and its meshing. The Bottom panel, shown exclusively with ITK-SNAP software, shows deformation fields and grids for clearer method differentiation. Deformed grids in the top-left panel (4th row) were generated using the Matplotlib library in Python. Top-right panel (2nd row) deformation fields map spatial dimensions $x,y,\mathrm{~and~}z$ to RGB color channels, respectively.}\label{fig4}
\end{figure}
\begin{figure}[!ht]
	\centering
	\includegraphics[width=0.86\textwidth,height=9.5cm]{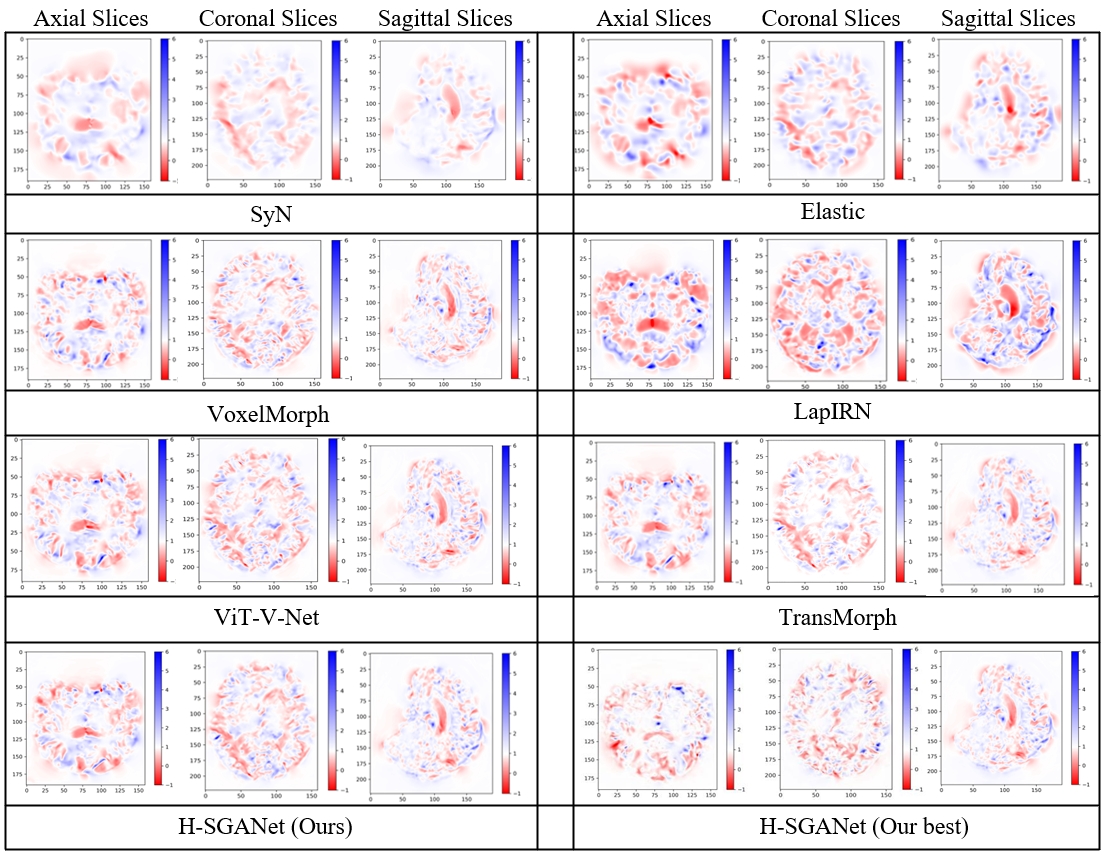}
	\caption{Quantitative analysis of Jacobian overlap. Color maps depict the Jacobian determinant of registration fields obtained through diverse methods on the OASIS dataset. Regions highlighted in bright red signify occurrences of voxel folding, indicating instances where the value is less than or equal to 0.}\label{fig5}
\end{figure}

\subsection{Qualitative Analysis} 
The qualitative results of our proposed framework, compared with six other state-of-the-art DMIR methods, are depicted in Fig. \ref{fig4}. The top-left, top-right, and bottom panels correspond to the three classic anatomical slices of OASIS brain MRI: sagittal, coronal, and axial. In the first row of each panel, the effectiveness of different models in aligning anatomical structures is showcased through the warped moving images. The detailed texture of the warped image in every slice, after registration by H-SGANet, is found to bear a stronger resemblance to the texture within a fixed image. The second and fourth rows exhibit the corresponding deformation fields and their grid representations. Our method consistently generates highly smooth deformation fields, even under intricate anatomical transformations. The third row illustrates the overlap between the warped and fixed images, highlighting overlapping regions in purple. Our method demonstrates a larger overlap, corresponding to a greater extent of the purple region, showcasing superior performance. In Fig. \ref{fig5}, the degree of folding in the estimated deformation fields is compared under different methods through visualization of the overlap quantification results. When the value is less than or equal to 0, voxel folding is indicated, displayed in dark red. A comparison of the extent of dark red regions reveals that our method exhibits relatively fewer folded voxels than other learning methods.

\subsection{Quantitative Analysis} 
\begin{figure}[!ht]
	\centering
	\includegraphics[width=0.805\textwidth]{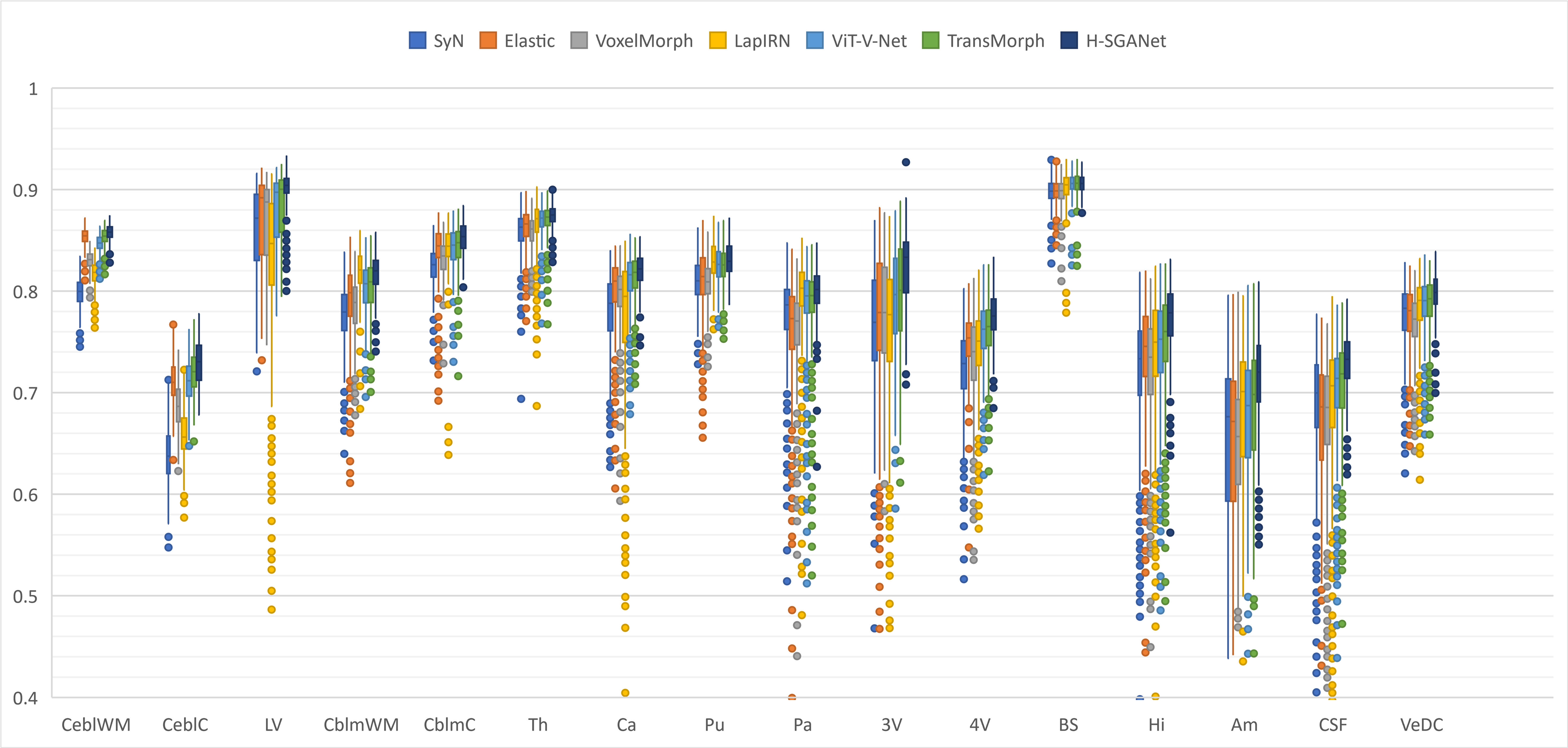}
	\caption{Boxplots depicting the average DSC of each anatomical structure in the OASIS test dataset for SyN, Elastic, VoxelMorph, LapIRN, ViT-V-Net, Transmorph, and our method. The left and right hemispheres of the brain are combined into one structure for visualization. The cerebral white matter((CeblWM), cerebral cortex (CeblC), lateral ventricle (LV), cerebellum cortex (CblmC), thalamus (Th), caudate (Ca), putamen (Pu),  pallidum (Pa), 3rd ventricle(3V), 4th ventricle (4V), brain stem (BS), hippocampus (Hi), amygdala (Am), CSF and ventral DC(VelDC) are included.}\label{fig6}
\end{figure}

\begin{figure}[!ht] 
	\centering  
	\subfigure[Parameters in each deep learning-based model. The
	values are expressed in the form of million parameters.]{
		\label{clr.sub.1}
		\includegraphics[width=0.47\linewidth]{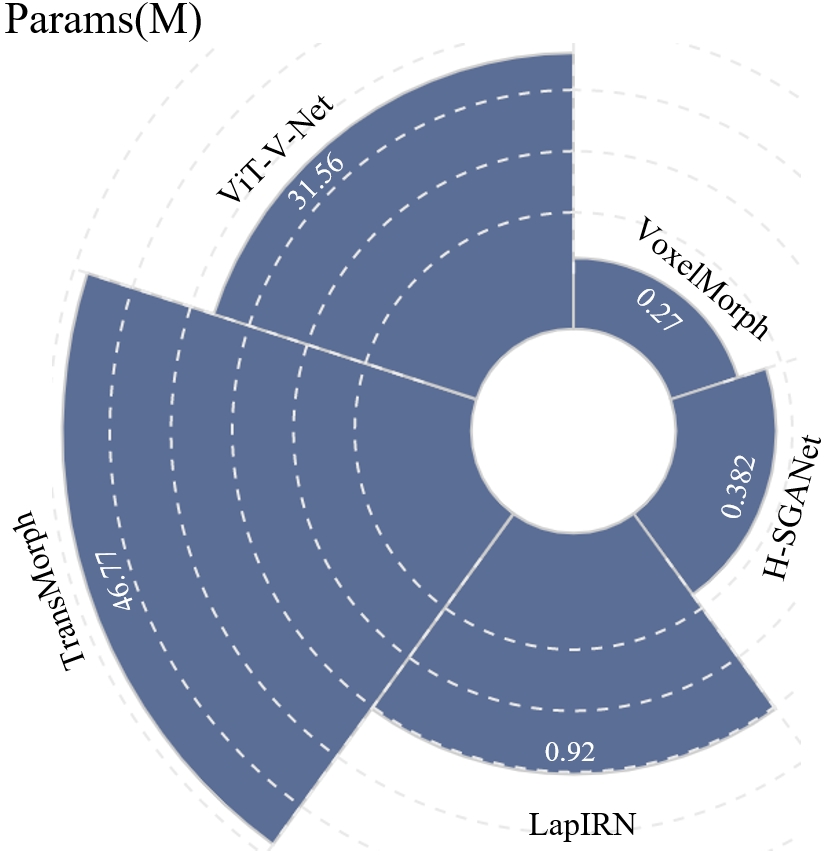}}\hspace{0.1cm} 
	\subfigure[Model FLOPs comparison. The unit is GMACs.]{
		\label{clr.sub.2}
		\includegraphics[width=0.47\linewidth]{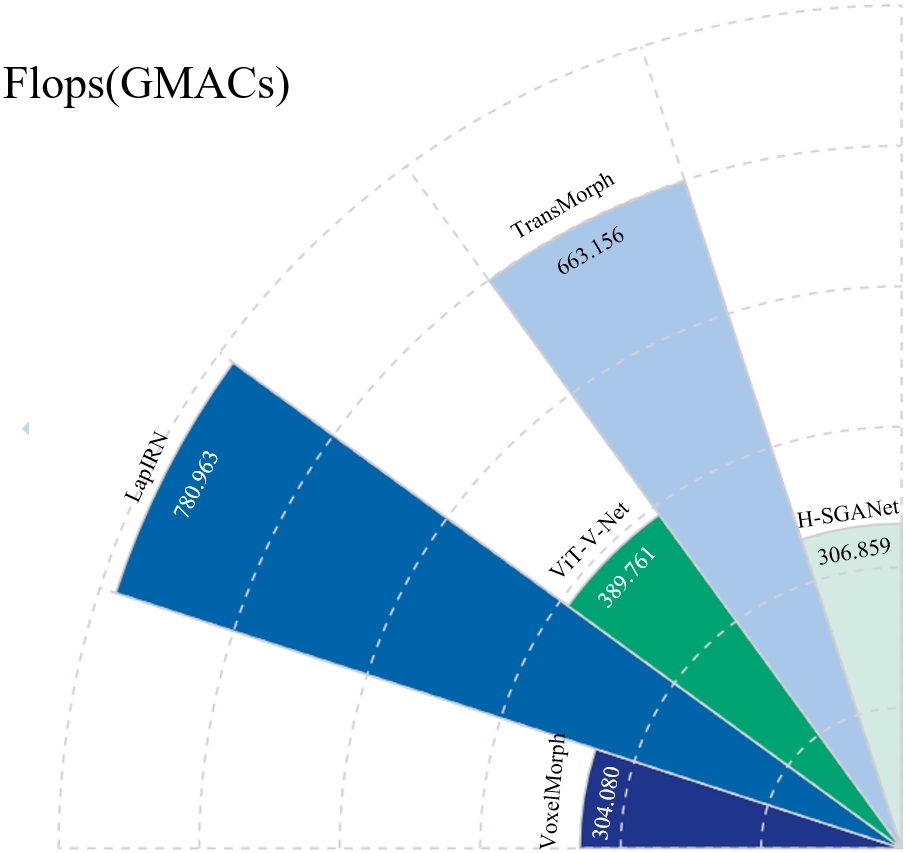}}
	\caption{Comparison of the volume of parameters and computational complexity.}
	\label{fig7}
\end{figure}

Table \ref{table3} shows the quantitative comparison of our method H-SGANet and other six DMIR methods for learn2reg OASIS brain MRI registration in the supervised setting. As shown in Table \ref{table3}, the deep learning-based methods exhibit more efficient computation and higher registration accuracy compared to SyN and Elastic. Our method, H-SGANet achieved the highest Dice score of 0.814. While the parameter count of our method is slightly higher than that of VoxelMorph, the computational effort is comparable, placing both metrics within the same order of magnitude (see Fig. \ref{fig7}). Importantly, the H-SGANet achieves a superior quantitative Dice score and a lower percentage of folded voxels (0.2\%). Note that even though ViT-V-Net and TransMorph having nearly a hundred times more trainable parameters (see Fig. \ref{fig7}), H-SGANet consistently outperformed all Transformer-based models, including the multi-level LapIRN method, by a margin of at least 0.1 in terms of DSC. This highlights the superiority of the hybrid ConvNet-ViG-Transformer model over each individual model. Cross-dataset validation results are presented in Table \ref{table4} for the proposed H-SGANet and other methods on LPBA40, where the model trained on the OASIS dataset is utilized to verify the generalization of networks. The LPBA40 dataset has more regions to be registered (i.e., 56 anatomical structures, see Table \ref{tab:2}), so the performance of each registration method is reduced to a certain extent. Compared with the multi-level cascaded LapIRN, the H-SGANet produces slightly lower Dice score while almost no voxel foldings (i.e., $0.08\% $ of $\vert J_{\phi}\vert \leq 0$). By leveraging the anatomical connections of the brain through SGA and capitalizing on the assistance of long-range dependencies provided by SSAFormer, our proposed method, H-SGANet, exhibits a substantial improvement in Dice score compared to other learning-based approaches. Fig. \ref{fig6} shows additional DSC values for a variety of anatomical structures using the proposed H-SGANet and baseline methods on the OASIS dataset. Notably, we incorporated the mirrored anatomical features of both the left and right hemispheres in our computations. The results presented in the boxplot show that the highest mean Dcie score of most anatomical structures is achieved by the proposed H-SGANet, especially on the cerebral cortex, putamen, and brain stem. This may be attributed to the expansion of the spatial connectivity range in the human brain, leading to an increased receptive field without a concurrent increase in computational effort. In summary, our H-SGANet network demonstrates superior performance in cross-dataset registration compared to ConvNet-based networks and Transformer-based networks, thus demonstrating its adaptability in different scenarios.
\begin{table}[!ht]
	\caption{Quantitative evaluation result of brain MRI registration on the OASIS dataset. DSC (the higher, the better) and percentage of $\vert J_{\phi}\vert \leq 0$ (NJD\%, the lower, the better); Time represents the mean duration for registering each pair of MR scans in seconds. Standard deviations are provided in brackets. Instances of bold font signify the best-performing results.}
	\label{table3}
	\centering
	\setlength{\tabcolsep}{1pt}
	\small
	\begin{tabular}{@{}ccccc@{}}
		\hline
		\multirow{2}*{Methods} & \  &OASIS & \ & \  \\
		\cline{2-5}
		\         & DSC          & $\%$ of $\vert J_{\phi}\vert \leq 0$       &Time(s) &  \makecell[c]{GPU memory\\ (training)} \\
		\hline
		Affine          & 0.611(0.019)   & -                              & -      & -     \\
		SyN             & 0.761(0.021)   & 0.00(0.00)         & 42.05  & -     \\
		Elastic          & 0.782(0.019)   & 0.29(0.05)          & 56.43  & -     \\
		VoxelMorph      & 0.779(0.007)   & 0.33(0.05)         & 0.35   & \textbf{9.970 GB}  \\
		LapIRN          & 0.783(0.002)    & 0.43(0.06)          & 0.88   & 15.48 GB     \\
		ViT-V-Net       & 0.797(0.004)   & 0.31(0.08)          & 0.39   & 10.09 GB     \\
		TransMorph      & 0.799(0.003)   & 0.21(0.05)         & 0.34   & 12.92 GB    \\
		H-SGANet(Ours)  & \textbf{0.814(0.005)}   & \textbf{0.20(0.04)}        & \textbf{0.26}   & 10.03 GB     \\
		\hline
	\end{tabular}
\end{table}
\begin{table}[!ht]
	\caption{Cross-dataset validation results for our approach and six other methods on the LPBA40 dataset were assessed. DSC (the higher, the better) and percentage of $\vert J_{\phi}\vert \leq 0$ (NJD\%, the lower, the better); Time represents the mean duration for registering each pair of MR scans in seconds. Standard deviations are provided in brackets. Instances of bold font signify the best-performing results.}
	\label{table4}
	\centering
	\setlength{\tabcolsep}{1pt}
	\small
	\begin{tabular}{@{}ccccc@{}}
		\hline
		\multirow{2}*{Methods} & \ &LPBA40  & \ &\  \\
		\cline{2-5}
		\         & DSC          & $\%$ of $\vert J_{\phi}\vert \leq 0$         &Time(s) &  \makecell[c]{GPU memory\\ (training)} \\
		\hline
		Affine          & 0.612(0.011)   & -                               & -      & -     \\
		SyN             & 0.698(0.015)   & 0.00(0.00)         & 28.98  & -     \\
		Elastic          & 0.699(0.016)   & 0.24(0.01)          & 50.62  & -     \\
		VoxelMorph      & 0.676(0.012)   & 0.29(0.06)          & 0.35   & \textbf{9.970 GB}  \\
		LapIRN          & \textbf{0.700(0.007)}    & 0.27(0.08)         & 0.88   & 15.48 GB     \\
		ViT-V-Net       & 0.648(0.007)   & 0.34(0.02)          & \textbf{0.23}   & 10.09 GB     \\
		TransMorph      & 0.661(0.006)   & 0.21(0.08)          & 0.35   & 12.92 GB    \\
		H-SGANet(Ours)  & 0.691(0.005)   & \textbf{0.08(0.04)}          & 0.28   & 10.03 GB      \\
		\hline
	\end{tabular}
\end{table}
\begin{table}[!ht]
	\caption{System-level comparison of diverse variants involving TransMorph and VoxelMorph designs on the OASIS datasets. Our method is evaluated against VoxelMorph, TransMorph, and several variants, including VoxelMorph + SGA, TransMorph + SGA, our method without Sparse Graph Attention (w/o SGA), and our method without Separable Self-Attention (w/o SSA). Bolded numbers indicate the highest scores.}
	\label{table5}
	\setlength{\tabcolsep}{0.5pt}
	\centering
	\small 
	\begin{tabular}{@{}ccccccc@{}}
		\hline
		\multirow{2}{*}{Method} & \multicolumn{3}{c}{OASIS}        & \multicolumn{3}{c}{LPBA40}       \\ \cline{2-7} 
		& DSC          & NJD \%     & Time & DSC          & NJD \%     & Time \\ \hline
		VoxelMorph              & 0.779(0.007) & 0.33(0.05) & 0.35 & 0.646(0.012) & 0.29(0.06) & 0.35 \\
		VM (+ SGA)              & 0.788(0.004) & 0.29(0.05) & 0.38 & 0.649(0.006) & 0.27(0.06) & 0.37 \\
		TransMorph              & 0.799(0.003) & 0.21(0.05) & 0.34 & 0661(0.006)  & 0.21(0.08) & 0.35 \\
		TM (+ SGA)     & \textbf{0.817(0.019)} & \textbf{0.20(0.03)} & 0.36          & 0.672(0.007)          & 0.20(0.02)          & 0.38          \\
		Ours (w/o SGA) & 0.796(0.006)          & 0.36(0.06)          & \textbf{0.28} & 0.656(0.013)          & 0.31(0.07)          & \textbf{0.27} \\
		Ours (w/o SSA)          & 0.807(0.012) & 0.25(0.04) & 0.33 & 0.682(0.011) & 0.17(0.02) & 0.30 \\ \hline
		Ours           & 0.814(0.003)          & 0.20(0.06)          & 0.29          & \textbf{0.697(0.019)} & \textbf{0.15(0.08)} & 0.28          \\ \hline
	\end{tabular}
\end{table}

\subsection{Model and Computational Complexity}
The fan chart depicted in Fig. \ref{fig7} provides a comprehensive comparison of parameter quantity and computational complexity among diverse deep learning-based registration models. All models considered in this analysis operate on a pair of MR images with dimensions set at 160×192×224. In this particular context, the H-SGANet model is introduced with a parameter count of 0.382 million, positioning it favorably when compared to other learning-based networks. It is worth noting that our approach prioritizes parameter efficiency, which is in line with the objective of achieving superior performance within limited parameter conditions. This is demonstrated by the GMACs of 306.859G associated with our method, indicating its effectiveness in registration tasks while utilizing parameters judiciously. 
\begin{figure}[!ht]
	\centering
	\includegraphics[width=0.9\textwidth]{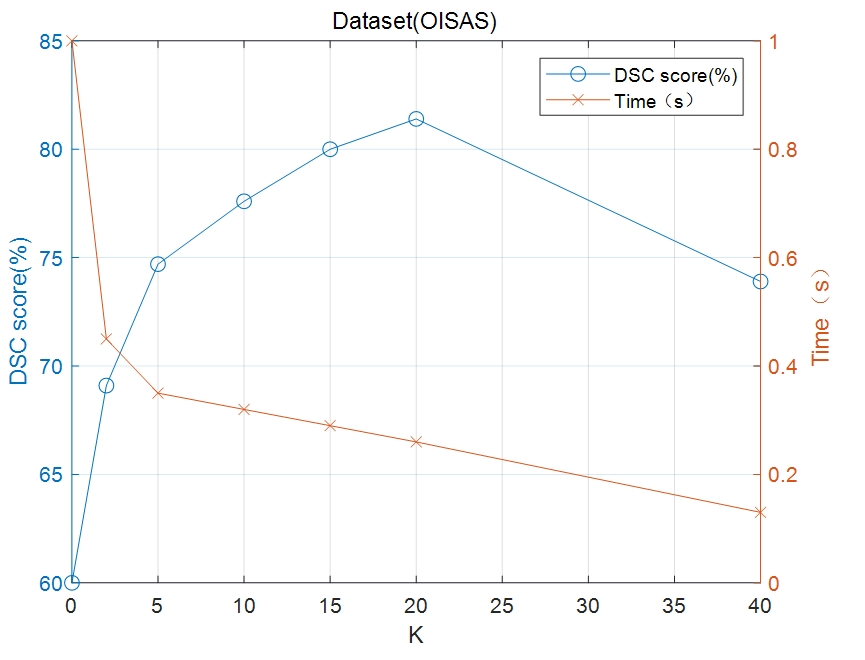}\\
	\caption{The optimum value of the K on OASIS and LPBA40 datasets.}\label{fig8}
\end{figure}
\begin{table}[htpb]
	\caption{The effects of modules in SGA for learn2reg OASIS brain MRI registration.}
	\label{table6}
	\centering
	\setlength{\tabcolsep}{1.5pt}
	\begin{tabular}{cccccc}
		\hline
		GraphConv    & \begin{tabular}{c}
			FC in Grapger\\ module
		\end{tabular} &\begin{tabular}{c}
	FFN\\ module 
	\end{tabular}    & Params(k) & FLOPs & DSC   \\ \hline
		$\usym{2713}$ & $\usym{2717}$        & $\usym{2717}$    & 214.4     & 172.1 & 0.731 \\
		$\usym{2713}$ & $\usym{2713}$         & $\usym{2717}$    & 162.6     & 172.6 & 0.799 \\
		$\usym{2713}$ & $\usym{2717}$        & $\usym{2713}$    & 284.6     & 159.7 & 0.801 \\
		$\usym{2713}$ & $\usym{2713}$        & $\usym{2713}$    & 262.5     & 160.2 & 0.805 \\ \hline
	\end{tabular}
\end{table}
\begin{table}[!ht]
	\centering
	\caption{Impact of distinct self-attention approaches on H-SGANet's performance assessed using the OASIS dataset. Inference time is averaged based on 30 repeated runs.}
	\label{table7}
	\begin{tabular}{p{4cm}cc}
		\hline
		Attention unit & Inference (sec/image) & Dice  \\ \hline
		Self-attention in Transformer & 0.329                & 0.798 \\
		Separable self-attention in SSAFormer     & 0.121                & 0.782 \\ \hline
	\end{tabular}
\end{table}

\subsection{Ablation study}
This section aims to identify the key factors contributing to the superior performance of our method. Our approach attains state-of-the-art results primarily by utilizing a hybrid ConvNet-ViG-Transformer architecture combined with SGA and SSAFormer fusion method for more accurate registration results. Then extensive ablation studies were conducted on the OASIS and LPBA40 datasets to analyze the impact of key modules in H-SGANet and interlayer stacking in SGA.

\textbf{Variant 1: SGA Module}  
We design a plug-and-play SGA module to better capture the connectivity in human brain space for better feature extraction. To figure out the strength of our method, we combine the SGA module with VoxelMorph (VM + SGA), and TransMorph (TM + SGA), and further evaluate our method’s performance without the module (replaced by mean pooling). Quantitative results in Table \ref{table5} show that the module benefits baseline models under two dataset scenes. TransMorph with the SGA module even outperforms ours in terms of performance on the OASIS dataset due to the benefits derived from larger model size and parameter quantity, enabling better adaptation to the data's complexity. However, this advantage may result in overfitting or increased computational resource demands. We contend that it is justifiable to compromise a certain degree of accuracy to conserve computational resources. As for the LPBA40 dataset, our module yielded only marginal performance improvements for the baseline methods, mainly due to the dataset's larger number of regions requiring alignment. Nevertheless, our method achieved the highest level of alignment accuracy despite these challenges. 

\textbf{Variant 2: The number of neighbors}
To adapt the ViG architecture for the registration task, FC layers are incorporated into the Grapher module of SGA and employed FFN blocks for feature transformation. Ablation studies assessing the effectiveness of these modules reveal that the direct use of graph convolution for registration yields suboptimal results. As depicted in Table \ref{table6}, the introduction of FC and FFN for increased feature transformations significantly enhances accuracy. Moreover, reducing $K$ in the SGA algorithm theoretically leads to increased runtime, as elucidated by analyzing the relationship between DSC scores and the average running time to register each pair of MR scans. As illustrated in Fig. \ref{fig8}, the algorithm containing only the SGA module at $K = 20$ achieves the highest DSC scores on both datasets.

\textbf{Variant 3: Use of SSAFormer}
To address the high computational resource overhead caused by MHA in Transformer when dealing with volumetric data, we propose the separable self-attention (SSA) module. In order to verify that our SSA module indeed has lower computational complexity and higher alignment accuracy compared to the multi-head self-attention (MHA) module in the traditional Transformer, we conducted experimental comparisons using SSAFormer and Transformer network on a simple U-shaped network structure, respectively. As shown in Table. \ref{table7}, When substituting the Multi-Head Attention (MHA) in the Transformer with a separable self-attention mechanism, we observe a threefold enhancement in inference speed while maintaining comparable performance on the Dice Similarity Coefficient (DSC) score. In terms of inference time, the learning-based model significantly surpasses traditional alignment methods by orders of magnitude. Notably, the Transformer-based method demonstrates an approximate threefold acceleration compared to the ConvNet-based method.

\section{Conclusions}
In this paper, we propose a novel unsupervised method
called H-SGANet for deformable medical image registration, which aims to emphasize salient brain network connections to enhance spatial understanding and capture inherent fixed connectivity relationships. Our SGA achieves promising registration performance for brain MRI volumes, resulting from capturing the anatomical connectivity within the internal architecture of the brain by introducing ViG and representing the image as a graph structure. Additionally, we propose SSA and SSAFormer, which exhibit linear complexity and effectively capture high-quality long-range dependencies, while also addressing the limitations of MHA in transformers. Extensive experiments were performed on two publicly available datasets, accompanied by detailed analyses, to demonstrate the state-of-the-art performance of our method in deformable image registration. We have further delved into the efficiency of our approach and evaluated the significance of its key components. Despite these advancements, our study has limitations that warrant further investigation. H-SGANet may still encounter challenges when dealing with images exhibiting extreme deformations and noise, suggesting the need for more robust model exploration in future work. Given this, we believe that this structure can also be applied to other visual tasks, such as image segmentation, image classification, medical image analysis, feature extraction, etc.

\section*{Acknowledgments}
This work was supported in part by the Fundamental Research Foundation of Shenzhen under Grant JCYJ20230808105705012.

 \bibliographystyle{elsarticle-num} 
 \bibliography{ref}

\begin{thebibliography}{10}
\expandafter\ifx\csname url\endcsname\relax
  \def\url#1{\texttt{#1}}\fi
\expandafter\ifx\csname urlprefix\endcsname\relax\def\urlprefix{URL }\fi
\expandafter\ifx\csname href\endcsname\relax
  \def\href#1#2{#2} \def\path#1{#1}\fi

\bibitem{deng2023interpretable}
X.~Deng, E.~Liu, S.~Li, Y.~Duan, M.~Xu, Interpretable multi-modal image registration network based on disentangled convolutional sparse coding, IEEE Transactions on Image Processing 32 (2023) 1078--1091.

\bibitem{avants2008symmetric}
B.~B. Avants, C.~L. Epstein, M.~Grossman, J.~C. Gee, Symmetric diffeomorphic image registration with cross-correlation: evaluating automated labeling of elderly and neurodegenerative brain, Medical image analysis 12~(1) (2008) 26--41.

\bibitem{modat2010fast}
M.~Modat, G.~R. Ridgway, Z.~A. Taylor, M.~Lehmann, J.~Barnes, D.~J. Hawkes, N.~C. Fox, S.~Ourselin, Fast free-form deformation using graphics processing units, Computer methods and programs in biomedicine 98~(3) (2010) 278--284.

\bibitem{sotiras2013deformable}
A.~Sotiras, C.~Davatzikos, N.~Paragios, Deformable medical image registration: A survey, IEEE transactions on medical imaging 32~(7) (2013) 1153--1190.

\bibitem{chen2023survey}
J.~Chen, Y.~Liu, S.~Wei, Z.~Bian, S.~Subramanian, A.~Carass, J.~L. Prince, Y.~Du, A survey on deep learning in medical image registration: New technologies, uncertainty, evaluation metrics, and beyond, arXiv preprint arXiv:2307.15615 (2023).

\bibitem{li2023d}
Z.~Li, Z.~Shang, J.~Liu, H.~Zhen, E.~Zhu, S.~Zhong, R.~N. Sturgess, Y.~Zhou, X.~Hu, X.~Zhao, et~al., D-lmbmap: a fully automated deep-learning pipeline for whole-brain profiling of neural circuitry, Nature Methods 20~(10) (2023) 1593--1604.

\bibitem{milletari2016v}
F.~Milletari, N.~Navab, S.-A. Ahmadi, V-net: Fully convolutional neural networks for volumetric medical image segmentation, in: 2016 fourth international conference on 3D vision (3DV), Ieee, 2016, pp. 565--571.

\bibitem{zheng2022multi}
Z.~Zheng, W.~Cao, Y.~Duan, G.~Cao, D.~Lian, Multi-strategy mutual learning network for deformable medical image registration, Neurocomputing 501 (2022) 102--112.

\bibitem{balakrishnan2019voxelmorph}
G.~Balakrishnan, A.~Zhao, M.~R. Sabuncu, J.~Guttag, A.~V. Dalca, Voxelmorph: a learning framework for deformable medical image registration, IEEE transactions on medical imaging 38~(8) (2019) 1788--1800.

\bibitem{chen2022transmorph}
J.~Chen, E.~C. Frey, Y.~He, W.~P. Segars, Y.~Li, Y.~Du, Transmorph: Transformer for unsupervised medical image registration, Medical image analysis 82 (2022) 102615.

\bibitem{ronneberger2015u}
O.~Ronneberger, P.~Fischer, T.~Brox, U-net: Convolutional networks for biomedical image segmentation, in: Medical image computing and computer-assisted intervention--MICCAI 2015: 18th international conference, Munich, Germany, October 5-9, 2015, proceedings, part III 18, Springer, 2015, pp. 234--241.

\bibitem{liu2021swin}
Z.~Liu, Y.~Lin, Y.~Cao, H.~Hu, Y.~Wei, Z.~Zhang, S.~Lin, B.~Guo, Swin transformer: Hierarchical vision transformer using shifted windows, in: Proceedings of the IEEE/CVF international conference on computer vision, 2021, pp. 10012--10022.

\bibitem{han2022vision}
K.~Han, Y.~Wang, J.~Guo, Y.~Tang, E.~Wu, Vision gnn: An image is worth graph of nodes, Advances in Neural Information Processing Systems 35 (2022) 8291--8303.

\bibitem{dosovitskiy2020image}
A.~Dosovitskiy, L.~Beyer, A.~Kolesnikov, D.~Weissenborn, X.~Zhai, T.~Unterthiner, M.~Dehghani, M.~Minderer, G.~Heigold, S.~Gelly, et~al., An image is worth 16x16 words: Transformers for image recognition at scale, arXiv preprint arXiv:2010.11929 (2020).

\bibitem{chen2021vit}
J.~Chen, Y.~He, E.~C. Frey, Y.~Li, Y.~Du, Vit-v-net: Vision transformer for unsupervised volumetric medical image registration, arXiv preprint arXiv:2104.06468 (2021).

\bibitem{zhang2021learning}
Y.~Zhang, Y.~Pei, H.~Zha, Learning dual transformer network for diffeomorphic registration, in: Medical Image Computing and Computer Assisted Intervention--MICCAI 2021: 24th International Conference, Strasbourg, France, September 27--October 1, 2021, Proceedings, Part IV 24, Springer, 2021, pp. 129--138.

\bibitem{SONG2022102612}
X.~Song, H.~Chao, X.~Xu, H.~Guo, S.~Xu, B.~Turkbey, B.~J. Wood, T.~Sanford, G.~Wang, P.~Yan, Cross-modal attention for multi-modal image registration, Medical Image Analysis 82 (2022) 102612.
\newblock \href {https://doi.org/https://doi.org/10.1016/j.media.2022.102612} {\path{doi:https://doi.org/10.1016/j.media.2022.102612}}.

\bibitem{shi2022xmorpher}
J.~Shi, Y.~He, Y.~Kong, J.-L. Coatrieux, H.~Shu, G.~Yang, S.~Li, Xmorpher: Full transformer for deformable medical image registration via cross attention, in: International Conference on Medical Image Computing and Computer-Assisted Intervention, Springer, 2022, pp. 217--226.

\bibitem{10158729}
Z.~Chen, Y.~Zheng, J.~C. Gee, Transmatch: A transformer-based multilevel dual-stream feature matching network for unsupervised deformable image registration, IEEE Transactions on Medical Imaging 43~(1) (2024) 15--27.
\newblock \href {https://doi.org/10.1109/TMI.2023.3288136} {\path{doi:10.1109/TMI.2023.3288136}}.

\bibitem{hou2021coordinate}
Q.~Hou, D.~Zhou, J.~Feng, Coordinate attention for efficient mobile network design, in: Proceedings of the IEEE/CVF conference on computer vision and pattern recognition, 2021, pp. 13713--13722.

\bibitem{zhang2021efficient}
Z.~Zhang, Y.~Wu, J.~Zhang, J.~Kwok, Efficient channel attention for deep convolutional neural networks, in: Proceedings of the AAAI Conference on Artificial Intelligence, 2021.

\bibitem{mehta2022separable}
S.~Mehta, M.~Rastegari, Separable self-attention for mobile vision transformers, arXiv preprint arXiv:2206.02680 (2022).

\bibitem{chen2021transunet}
J.~Chen, Y.~Lu, Q.~Yu, X.~Luo, E.~Adeli, Y.~Wang, L.~Lu, A.~L. Yuille, Y.~Zhou, Transunet: Transformers make strong encoders for medical image segmentation, arXiv preprint arXiv:2102.04306 (2021).

\bibitem{zu2021van}
Z.~Zu, G.~Zhang, Y.~Peng, Z.~Ye, C.~Shen, Van: Voting and attention based network for unsupervised medical image registration, in: PRICAI 2021: Trends in Artificial Intelligence: 18th Pacific Rim International Conference on Artificial Intelligence, PRICAI 2021, Hanoi, Vietnam, November 8--12, 2021, Proceedings, Part I 18, Springer, 2021, pp. 382--393.

\bibitem{woo2018cbam}
S.~Woo, J.~Park, J.-Y. Lee, I.~S. Kweon, Cbam: Convolutional block attention module, in: Proceedings of the European conference on computer vision (ECCV), 2018, pp. 3--19.

\bibitem{chen2023dusfe}
X.~Chen, B.~Zhou, H.~Xie, X.~Guo, J.~Zhang, J.~S. Duncan, E.~J. Miller, A.~J. Sinusas, J.~A. Onofrey, C.~Liu, Dusfe: Dual-channel squeeze-fusion-excitation co-attention for cross-modality registration of cardiac spect and ct, Medical Image Analysis 88 (2023) 102840.

\bibitem{hu2018squeeze}
J.~Hu, L.~Shen, G.~Sun, Squeeze-and-excitation networks, in: Proceedings of the IEEE conference on computer vision and pattern recognition, 2018, pp. 7132--7141.

\bibitem{gori2005new}
M.~Gori, G.~Monfardini, F.~Scarselli, A new model for learning in graph domains, in: Proceedings. 2005 IEEE International Joint Conference on Neural Networks, 2005., Vol.~2, IEEE, 2005, pp. 729--734.

\bibitem{scarselli2008graph}
F.~Scarselli, M.~Gori, A.~C. Tsoi, M.~Hagenbuchner, G.~Monfardini, The graph neural network model, IEEE transactions on neural networks 20~(1) (2008) 61--80.

\bibitem{niepert2016learning}
M.~Niepert, M.~Ahmed, K.~Kutzkov, Learning convolutional neural networks for graphs, in: International conference on machine learning, PMLR, 2016, pp. 2014--2023.

\bibitem{kipf2016semi}
T.~N. Kipf, M.~Welling, Semi-supervised classification with graph convolutional networks, arXiv preprint arXiv:1609.02907 (2016).

\bibitem{defferrard2016convolutional}
M.~Defferrard, X.~Bresson, P.~Vandergheynst, Convolutional neural networks on graphs with fast localized spectral filtering, Advances in neural information processing systems 29 (2016).

\bibitem{hamilton2017inductive}
W.~Hamilton, Z.~Ying, J.~Leskovec, Inductive representation learning on large graphs, Advances in neural information processing systems 30 (2017).

\bibitem{ying2018graph}
R.~Ying, R.~He, K.~Chen, P.~Eksombatchai, W.~L. Hamilton, J.~Leskovec, Graph convolutional neural networks for web-scale recommender systems, in: Proceedings of the 24th ACM SIGKDD international conference on knowledge discovery \& data mining, 2018, pp. 974--983.

\bibitem{kojima2020kgcn}
R.~Kojima, S.~Ishida, M.~Ohta, H.~Iwata, T.~Honma, Y.~Okuno, kgcn: a graph-based deep learning framework for chemical structures, Journal of Cheminformatics 12 (2020) 1--10.

\bibitem{touvron2021training}
H.~Touvron, M.~Cord, M.~Douze, F.~Massa, A.~Sablayrolles, H.~J{\'e}gou, Training data-efficient image transformers \& distillation through attention, in: International conference on machine learning, PMLR, 2021, pp. 10347--10357.

\bibitem{yang2022graformerdir}
T.~Yang, X.~Bai, X.~Cui, Y.~Gong, L.~Li, Graformerdir: Graph convolution transformer for deformable image registration, Computers in Biology and Medicine 147 (2022) 105799.

\bibitem{wang2023diegraph}
L.~Wang, Z.~Yan, W.~Cao, J.~Ji, Diegraph: dual-branch information exchange graph convolutional network for deformable medical image registration, Neural Computing and Applications 35~(32) (2023) 23631--23647.

\bibitem{li2019deepgcns}
G.~Li, M.~Muller, A.~Thabet, B.~Ghanem, Deepgcns: Can gcns go as deep as cnns?, in: Proceedings of the IEEE/CVF international conference on computer vision, 2019, pp. 9267--9276.

\bibitem{yu2022metaformer}
W.~Yu, M.~Luo, P.~Zhou, C.~Si, Y.~Zhou, X.~Wang, J.~Feng, S.~Yan, Metaformer is actually what you need for vision, in: Proceedings of the IEEE/CVF conference on computer vision and pattern recognition, 2022, pp. 10819--10829.

\bibitem{he2016deep}
K.~He, X.~Zhang, S.~Ren, J.~Sun, Deep residual learning for image recognition, in: Proceedings of the IEEE conference on computer vision and pattern recognition, 2016, pp. 770--778.

\bibitem{bajcsy1989multiresolution}
R.~Bajcsy, S.~Kova{\v{c}}i{\v{c}}, Multiresolution elastic matching, Computer vision, graphics, and image processing 46~(1) (1989) 1--21.

\bibitem{mok2020large}
T.~C. Mok, A.~C. Chung, Large deformation diffeomorphic image registration with laplacian pyramid networks, in: Medical Image Computing and Computer Assisted Intervention--MICCAI 2020: 23rd International Conference, Lima, Peru, October 4--8, 2020, Proceedings, Part III 23, Springer, 2020, pp. 211--221.

\end{thebibliography}
\end{document}